%% file: main.tex
\pdfoutput=1
\documentclass[11pt]{article}

\usepackage[utf8]{inputenc}
\usepackage[a4paper,margin=1in]{geometry}
\usepackage{graphicx}
\usepackage{booktabs}
\usepackage{amsmath}
\usepackage{amssymb}
\usepackage{nicefrac}
\usepackage{microtype}
\usepackage{xcolor}
\usepackage{subcaption}
\usepackage{multirow}
\usepackage{enumitem}
\usepackage{wrapfig}
\usepackage{url}
\usepackage[colorlinks=true,linkcolor=blue,citecolor=blue,urlcolor=blue]{hyperref}

\title{Empowering Feed-Forward Reconstruction Models with Metric Scale via Satellite Images}
\author{%
  \small
  Xianghui Ze\textsuperscript{1}, 
  Yongjian Luo\textsuperscript{2}, 
  Mengjun Chao\textsuperscript{2}, 
  Zhenbo Song\textsuperscript{1},  
  Jianfeng Lu\textsuperscript{1},
  Yujiao Shi\textsuperscript{2}\\[0.5em]
  \textsuperscript{1}Nanjing University of Science and Technology, 
  \textsuperscript{2}ShanghaiTech University\\[0.3em]
  \{zexh,songzb,lujf\}@njust.edu.cn, 
  \{luoyj2024,v-chaomj,shiyj2\}@shanghaitech.edu.cn
}
\date{}
\vspace{-1em}
\begin{document}
\maketitle

\begin{abstract}
Feed-forward 3D reconstruction models have recently shown strong generalization across diverse scenes, yet most of them recover geometry only up to an unknown global scale. This scale ambiguity limits their use in applications that require metric understanding of the environment. Existing metric reconstruction methods commonly rely on large-scale metric annotations or accurate camera calibration, both of which are costly or unreliable in many real-world settings. We propose a satellite-guided framework for resolving scale ambiguity in feed-forward 3D reconstruction. The key idea is to use readily available satellite imagery as a global metric reference. Given a coarse camera pose, our method retrieves a local satellite patch and integrates it with a feed-forward reconstruction backbone through bidirectional cross-view interaction. By enforcing consistency between the reconstructed scene and the satellite reference, the model infers absolute scale, refines scene geometry, and estimates camera pose in a metric coordinate frame. Experiments on KITTI, nuScenes, and Oxford RobotCar show consistent improvements in metric depth estimation, multi-view point-cloud reconstruction, and cross-view camera localization, while preserving strong generalization across datasets and geographic regions.
\vspace{-1em}
\end{abstract}

\input{secs/Introduction}
\input{secs/RelatedWork}
\input{secs/Method}
\input{secs/Experiments}

\section{Conclusion}
We address the long-standing scale ambiguity in feed-forward 3D reconstruction by using satellite imagery as a global metric reference. Rather than depending on large-scale metric annotations or accurately calibrated cameras, the proposed framework exploits widely available satellite observations to provide complementary geometric cues. This enables the model to recover metric scene geometry and absolute camera pose while preserving the generalization capability of scale-agnostic foundation models.

Extensive experiments demonstrate consistent gains in metric depth estimation, point-cloud reconstruction, and camera localization across datasets and geographic regions. These results suggest that satellite imagery is a practical and scalable source of metric information for visual geometry learning. Future work may incorporate richer geometric priors, such as building footprints or floor plans, to extend metric reconstruction to a broader range of indoor and outdoor environments.

\bibliographystyle{unsrt}
\bibliography{main}


\end{document}

%% file: secs/Introduction.tex
\section{Introduction}

Visual geometry reconstruction is a fundamental problem in computer vision. Reliable reconstruction is critical for downstream applications such as augmented reality~\cite{cao2020accurate}, autonomous driving~\cite{liao2025learning}, and embodied AI~\cite{duan2022survey}. Compared with traditional pipelines such as Structure-from-Motion (SfM)~\cite{ozyecsil2017survey} and Multi-View Stereo (MVS)~\cite{furukawa2015multi}, recent feed-forward 3D reconstruction methods~\cite{murai2025mast3r,wang2025vggt,wang2025pi} enable efficient end-to-end inference without iterative optimization and show strong robustness across diverse scenes. However, most feed-forward methods recover only scale-agnostic geometry: the absolute metric scale of the scene remains ambiguous. This limitation restricts their use in real-world tasks that require metric understanding of 3D space.

\begin{figure}[t]
    \centering
    \includegraphics[width=1\textwidth]{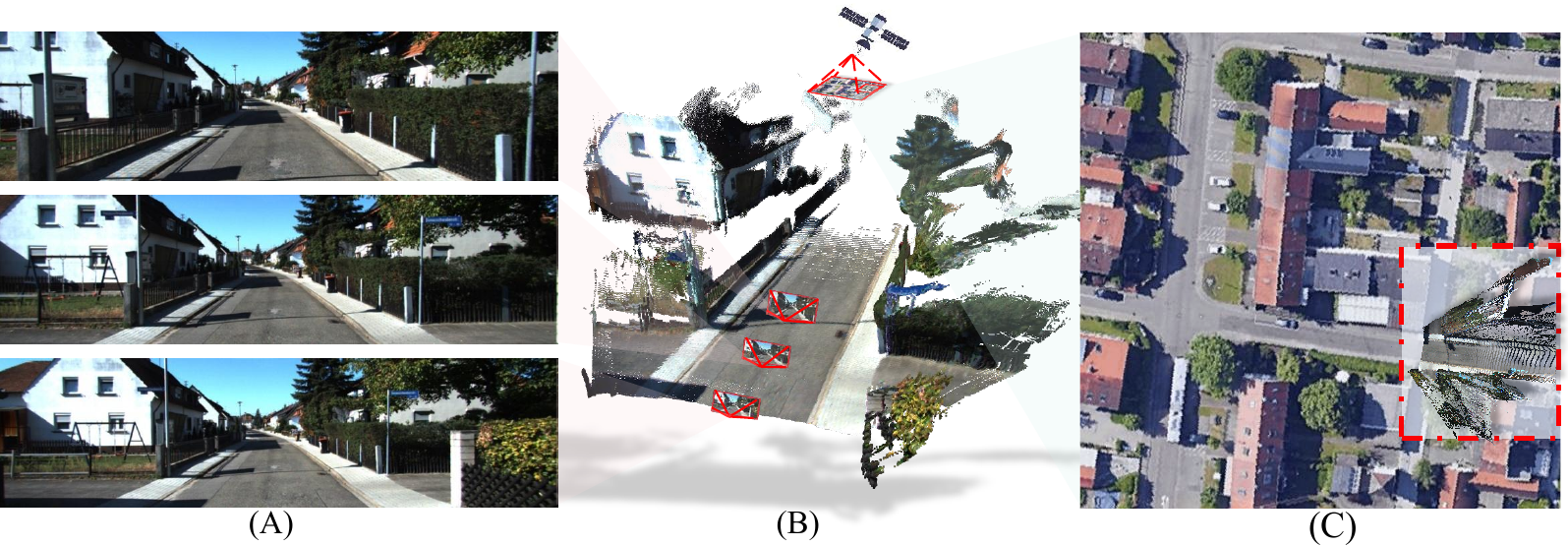}
    \caption{By incorporating satellite patches retrieved from coarse GPS signals, our method resolves the scale ambiguity of foundation-model-based 3D estimation and achieves metric scene reconstruction together with absolute camera pose estimation. In (C), the red box illustrates the projection of the reconstructed metric scene onto the satellite image using the estimated absolute camera pose, demonstrating accurate geometric alignment.}
    \vspace{-1em}
    \label{fig:pipeline}
\end{figure}

Recent works~\cite{piccinelli2025unidepthv2,wang2025moge,lin2025depth} have recognized the importance of metric depth and attempted to mitigate scale ambiguity by training models on large-scale datasets with metric depth annotations. Through such supervision, models can learn object-size priors and produce approximate metric predictions. While this strategy improves metric estimation in familiar environments, it requires massive annotated datasets and often generalizes poorly to unseen scenes or object categories. Another line of work, represented by feed-forward reconstruction frameworks~\cite{murai2025mast3r,wang2025pi}, focuses on scale-agnostic geometric representations. By avoiding explicit metric supervision, these methods achieve stronger robustness across scenes, but the lack of absolute scale fundamentally limits their applicability. Several approaches~\cite{lin2025depth,liu2024metricdepth,ashida2024resolving} introduce additional cues, such as camera extrinsic parameters, to inject metric information into reconstruction. Although effective under controlled settings, accurate camera parameters are often unavailable or unreliable in practice. These challenges stem from the inherent scale ambiguity of visual perception: image observations alone do not uniquely determine absolute scene scale, and multiple 3D reconstructions differing by a global scale factor can explain the same visual evidence. Recovering metric geometry therefore requires external references or additional geometric constraints.

We propose to introduce metric scale information into feed-forward 3D reconstruction by leveraging satellite imagery as an external reference. Satellite images are widely available and encode the physical layout of outdoor environments in a metric coordinate system. They therefore provide complementary cues for resolving scale ambiguity without requiring accurate camera calibration or dense metric annotations.

Directly using satellite imagery for geometric reasoning typically requires precise GPS localization of the ground camera to retrieve the corresponding satellite region. However, precise GPS measurements are not always available or reliable in real-world scenarios~\cite{xia2023convolutional}. We relax this requirement and assume only a coarse GPS estimate, which is sufficient to retrieve a local satellite patch covering the surrounding area. From this coarse initialization, our framework jointly aligns ground-view observations with the satellite patch to infer absolute camera pose and metric scene geometry. This design allows the model to exploit satellite imagery as a global spatial reference and enables unified estimation of camera localization and metric 3D reconstruction.

Specifically, our pipeline first retrieves a local satellite patch from a satellite image database using the coarse pose of a ground camera. Built upon a feed-forward geometry foundation model trained on large-scale datasets, the framework preserves strong generalization while introducing satellite imagery as an external metric reference. To exploit this reference, we design bidirectional cross-view interaction together with a scene-to-satellite alignment loss, explicitly encouraging consistency between the ground-view reconstruction branch and the satellite-reference branch. By enforcing cross-view alignment between ground-view geometry and satellite observations, the model learns to infer global scale and spatial alignment, enabling metric-scale 3D reconstruction.

In summary, our contributions are as follows:
\begin{itemize}
\item We introduce satellite imagery as a practical source of metric cues for feed-forward 3D reconstruction, enabling metric geometry estimation from coarse camera poses.
\item We propose a unified framework that jointly estimates metric scene geometry and absolute camera pose, making feed-forward reconstruction directly usable in downstream metric applications.
\item We design a scene-to-satellite alignment objective that enforces consistency between reconstructed scene representations and corresponding satellite imagery.
\item We conduct extensive experiments on multiple datasets, demonstrating the effectiveness and generalization capability of the proposed approach.
\end{itemize}

%% file: secs/RelatedWork.tex
\section{Related Work}

\textbf{Feed-forward 3D reconstruction.} Feed-forward 3D reconstruction has recently emerged as an efficient alternative to traditional pipelines based on structure-from-motion~\cite{schonberger2016pixelwise,iglhaut2019structure} and multi-view stereo~\cite{hartley2003multiple,goesele2006multi}. These methods infer 3D structure directly from images through a single neural-network forward pass. DUSt3R~\cite{wang2024dust3r} and its successors~\cite{leroy2024grounding,smart2024splatt3r,wang20253d} formulate geometric 3D reasoning as a feed-forward visual prediction problem, enabling direct reconstruction from image pairs without conventional optimization. Fast3R~\cite{yang2025fast3r} extends feed-forward reconstruction to large-scale settings and reconstructs scenes from more than one thousand images in a single forward pass. Methods such as FLARE~\cite{zhang2025flare} further integrate geometry, appearance, and camera estimation into a unified feed-forward framework under uncalibrated and sparse-view conditions. More recently, VGGT~\cite{wang2025vggt,wang2025faster} and $\pi^3$~\cite{wang2025pi} introduce visual geometric priors within Transformer-based architectures, enabling global reasoning over multi-view inputs. These methods show impressive generalization across scenes and datasets. Nevertheless, to preserve such generalization, most feed-forward approaches primarily recover relative geometry and avoid explicit metric-scale estimation. Although recent metric depth models~\cite{lin2025depth} attempt to encode metric information through large-scale training, their performance can be limited by dataset-specific priors and may rely on camera information that is not always reliable in real-world settings.

\textbf{Satellite-image-based camera pose estimation.} Satellite-image-based camera pose estimation localizes a ground camera by comparing ground-level observations with satellite imagery. Retrieval-based methods~\cite{zhu2021vigor,vivanco2023geoclip,vyas2022gama,shi2022cvlnet,shi2020looking} identify the most similar satellite patches from an image database, but their accuracy and efficiency are often constrained by database size. Other methods refine an initial GPS estimate. One line of work~\cite{shi2022beyond,shi2023boosting} iteratively optimizes the alignment between ground and satellite images, while another line~\cite{xia2023convolutional,lentsch2023slicematch,tong2025geodistill} samples candidate poses from satellite imagery and selects the best one by evaluating ground-image consistency. A third class of methods~\cite{song2023learning,wang2023fine,xia2025fg} establishes cross-view correspondences and estimates relative pose from matched points. These approaches demonstrate that satellite imagery provides a low-cost and widely available geographic reference. Inspired by this observation, we incorporate satellite imagery into feed-forward 3D reconstruction, not only to localize the camera but also to recover metric-scale scene geometry.

%% file: secs/Method.tex
\section{Method}

\begin{figure}[t]
    \centering
    \includegraphics[width=1\textwidth]{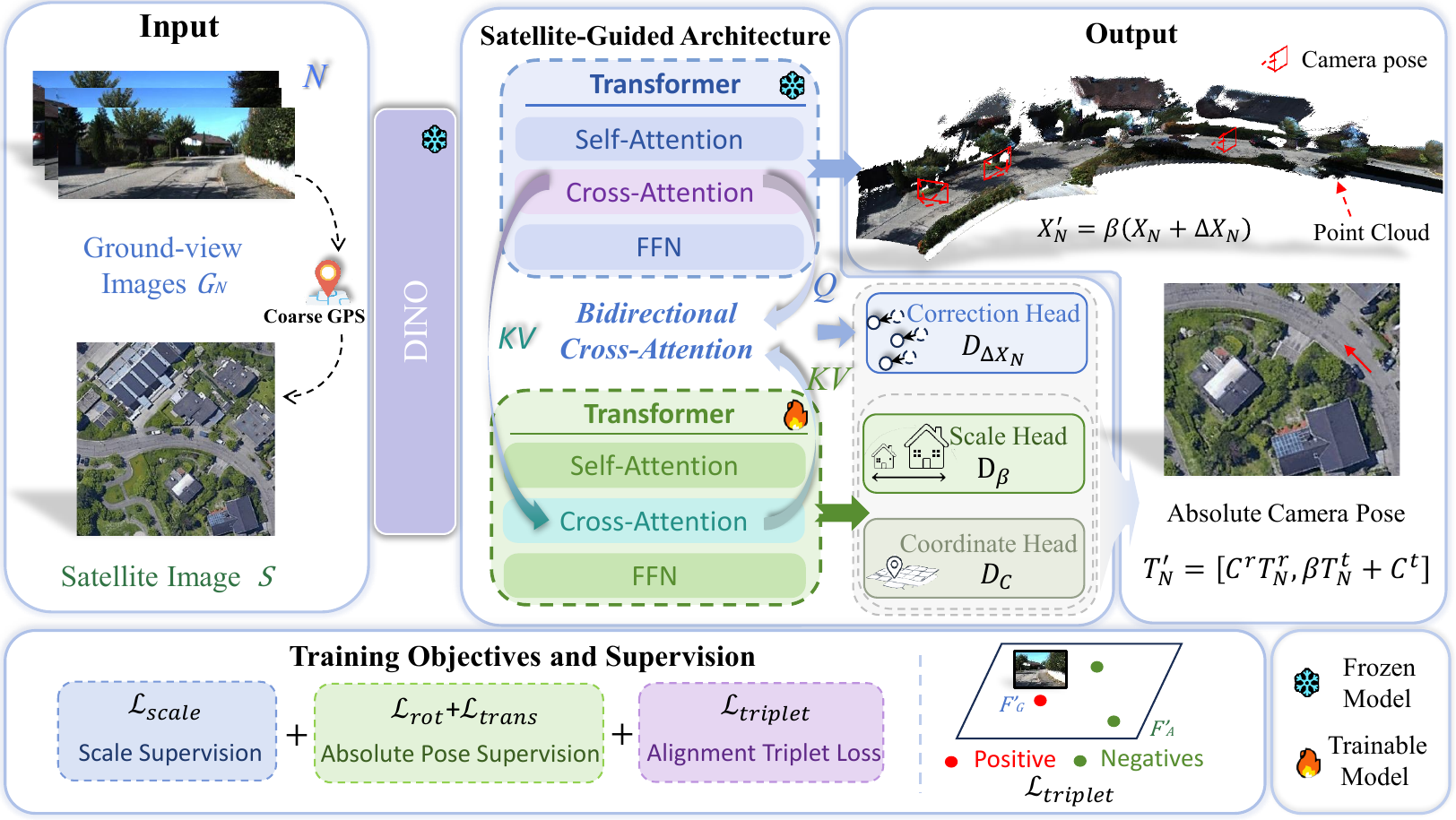}
    \caption{Overview of the proposed framework. Given a coarse ground-view pose, our method retrieves a corresponding satellite patch as a global metric reference. Through bidirectional interaction between satellite and ground-view branches, the network estimates a global scale factor $\beta$, a coordinate offset $C$, and a geometric correction term $\Delta X_N$, enabling metric scene reconstruction and absolute camera pose estimation in the satellite coordinate frame.}
    \label{fig:pipeline_method}
    \vspace{-1em}
\end{figure}

\subsection{Problem Formulation}
\textbf{Preliminary.} Recent feed-forward 3D reconstruction methods have demonstrated strong performance across diverse datasets and scenarios. Given a single image or a sequence of $N$ images $G_N$, a base feed-forward model $\phi(\cdot)$ predicts camera motion and scene geometry as
\begin{equation}
(T_N, X_N) = \phi(G_N),
\end{equation}
where $T_N \in \mathbb{R}^{N \times 4 \times 4}$ denotes the estimated camera poses, and $X_N \in \mathbb{R}^{N \times 3}$ represents the reconstructed 3D points. The pose of the $i$-th camera is written as $T_i = [T_i^r, T_i^t]$, where $T_i^r \in \mathrm{SO}(3)$ and $T_i^t \in \mathbb{R}^3$ denote rotation and translation, respectively.

These feed-forward formulations are inherently scale-agnostic: without external metric cues, both camera motion and scene geometry are defined only up to an arbitrary global scale. While such representations generalize well, the absence of absolute scale limits their applicability in tasks requiring metric reasoning.

\textbf{Our approach.} Our goal is to recover metric 3D geometry and accurate vehicle localization using only coarse GPS position estimates. We retrieve satellite imagery according to the coarse GPS measurements and use it as an external metric reference. Given the scale-agnostic output of the base model, our method predicts a scale factor $\beta$, a coordinate offset $C$, and a geometric correction $\Delta X_N$:
\begin{equation}
{T}_N^{'} = [C^r T_N^r,\; \beta T_N^t + C^t], \quad
{X}_N^{'} = \beta (X_N + \Delta X_N),
\label{eq:pre}
\end{equation}
where $\beta$ denotes the global scale factor, $C=(C^r,C^t)$ aligns the reconstruction to the satellite coordinate system, and $\Delta X_N$ refines the reconstructed point cloud.

\subsection{Satellite-Augmented Metric Scale Estimation}
Our framework injects satellite imagery into a feed-forward 3D reconstruction pipeline to resolve scale ambiguity while preserving generalization. We adopt $\pi^3$ as a representative backbone, although the design is applicable to other feed-forward reconstruction models.

To incorporate satellite information, we freeze the pretrained base model $\phi$ and clone it into a trainable satellite branch. The frozen branch processes ground-view images and preserves the original reconstruction capability, while the satellite branch processes satellite imagery and learns to inject metric cues in a way that is consistent with the base predictions.

Effectively leveraging satellite imagery requires reliable cross-view correspondence. In practice, satellite and ground-view images are captured from drastically different viewpoints and often at different times, causing discrepancies due to seasonal changes, illumination variation, and dynamic objects. We address this challenge by modeling mutual dependencies between satellite and ground-view features through cross-attention. This design allows the network to selectively fuse consistent information while suppressing mismatched regions, and it provides a mechanism for inferring physical alignment, including global scale and coordinate offsets.

Formally, let $F_S \in \mathbb{R}^{M \times d}$ and $F_G \in \mathbb{R}^{N \times d}$ denote features extracted from the satellite image and the ground-view image. We perform bidirectional cross-attention between these two feature sets. In the first direction, satellite features act as queries, while ground-view features serve as keys and values:
\begin{equation}
\tilde{F}_S = \mathrm{softmax}\!\left( \frac{Q_S K_G^\top}{\sqrt{d}} \right) V_G,
\end{equation}
where $Q_S = W_Q^S F_S$, $K_G = W_K^S F_G$, and $V_G = W_V^S F_G$ are the projected queries, keys, and values. This operation allows each satellite feature to aggregate information from geometrically and semantically consistent ground-view regions, enabling estimation of the global scale and coordinate offsets with respect to the satellite reference frame.

In the reverse direction, ground-view features serve as queries, while satellite features act as keys and values:
\begin{equation}
\tilde{F}_G = \mathrm{softmax}\!\left( \frac{Q_G K_S^\top}{\sqrt{d}} \right) V_S,
\end{equation}
where $Q_G = W_Q^G F_G$, $K_S = W_K^G F_S$, and $V_S = W_V^G F_S$. This interaction injects metric and structural cues from satellite imagery into the ground-view representation, allowing the model to refine the scene geometry reconstructed from ground-view images alone.

Based on the cross-view interaction features, we use three task-specific prediction heads to estimate global scale, coordinate offset, and geometric correction:
\begin{equation}
\beta = D_{\beta}(\tilde{F}_S),\quad
C = D_C(\tilde{F}_S),\quad
\Delta X_N = D_{\Delta X_N}(\tilde{F}_G).
\end{equation}
The scale head $D_{\beta}$ applies global average pooling followed by a linear layer to output the scalar $\beta$. The coordinate head $D_C$ uses a multilayer perceptron to regress the rotation and translation parameters $C=[C^r,C^t]$. The geometry correction head $D_{\Delta X_N}$ predicts the point-cloud refinement $\Delta X_N$, compensating for reconstruction errors and improving geometric accuracy.

\subsection{Training Objectives and Supervision}
\textbf{Scale factor supervision.} To recover metric scene geometry, we supervise the global scale factor that aligns the scale-agnostic point map $X_N$ with the ground-truth metric point cloud $\hat{X}_N$. Given the predicted and ground-truth point clouds, we compute the optimal scale factor $\hat{\beta}$ by minimizing their $\ell_1$ distance:
\begin{equation}
\hat{\beta} = \arg\min_{\beta}\Vert\hat{X}_N - \beta X_N\Vert_1.
\end{equation}
Unlike prior methods~\cite{wang2025vggt,wang2025pi} that resolve scale ambiguity by post-hoc alignment to ground truth, our method predicts $\beta$ during inference. We define the scale supervision loss as
\begin{equation}
\mathcal{L}_{\mathrm{scale}}
= \frac{(\beta - \hat{\beta})^2}{\hat{\beta}^{\,2} + \sigma},
\end{equation}
where $\sigma$ is a small constant for numerical stability. Normalization by the squared optimal scale reduces sensitivity to scene-scale variation and encourages accurate relative scale estimation.

\textbf{Absolute pose supervision.} By aligning the reconstructed scene with satellite imagery, we localize the camera in the satellite reference frame and recover its absolute pose. Following Eq.~\ref{eq:pre}, the absolute camera pose is constructed as $T_N^{'}= [C^r T_N^r,\; \beta T_N^t + C^t]$. Let $\hat{P}_N = [\hat{P}_N^r, \hat{P}_N^t]$ denote the ground-truth absolute camera pose in the satellite coordinate frame. We define rotation and translation losses as
\begin{equation}
\mathcal{L}_{\mathrm{rot}}
= \arccos\!\left(
\frac{\operatorname{Tr}\!\left((C^rT_N^r)^{\top}\hat{P}_N^r\right) - 1}{2}
\right), \quad
\mathcal{L}_{\mathrm{trans}} = \mathcal{H}(\hat{\beta} T_N^t + C^t - \hat{P}_N^t),
\end{equation}
where $\mathcal{H}(\cdot)$ denotes the Huber loss. To decouple pose learning from scale prediction during training, we substitute the ground-truth scale $\hat{\beta}$ for the predicted scale in the translation loss.

\textbf{Satellite alignment consistency supervision.} To encourage effective use of satellite imagery for metric reasoning, we explicitly supervise alignment consistency between the ground-view image and its corresponding satellite image. The core idea is to enforce geometric and semantic consistency between the reconstructed scene, expressed in a bird's-eye-view (BEV) representation, and the satellite reference. Given the geometry estimated by the foundation model and the recovered scale and global coordinate offset, we project the ground-view feature into BEV space:
\begin{equation}
f_G^{\text{BEV}} = P\!\left(f_G,\;\beta (X_N + \Delta X_N)\right),
\label{Eq:project}
\end{equation}
where $P(\cdot)$ is a differentiable projection operator~\cite{shi2022beyond}, $f_G$ denotes the ground-view feature, $\Delta X_N$ is the learnable geometric refinement, and $\beta$ controls the metric scale of the scene geometry.

We then supervise the similarity between the BEV-projected representation $f_G^{\text{BEV}}$ and the corresponding satellite feature $f_S$. Specifically, we extract visual features from both satellite and ground-view images using a pretrained DINO encoder. A multi-resolution aggregation module harmonizes the two views, yielding ground-view embeddings $f_G$ and satellite embeddings $f_S$. The ground-view features are projected into BEV space according to Eq.~\ref{Eq:project}.

We use a triplet-style loss to supervise similarity between BEV-projected ground features and satellite features. For each input ground-view image, the ground-truth camera location $(\hat{u}_i, \hat{v}_i)$ on the satellite map is treated as the positive sample, while other satellite locations $(u,v)$ are treated as negatives:
\begin{equation}
\mathcal{L}_{\text{triplet}} =
\frac{1}{HW - 1}
\sum_{\substack{(u,v) \neq (\hat{u}_i, \hat{v}_i)}}
\log\!\left(
1 +
\exp\!\left(
\alpha
\left(
\mathcal{C}_i(\hat{u}_i, \hat{v}_i)
-
\mathcal{C}_i(u, v)
\right)
\right)
\right),
\end{equation}
where $H$ and $W$ are the height and width of the satellite feature map. The similarity score $\mathcal{C}_i(u,v)$ measures the correspondence between the BEV-projected ground-view feature and the satellite feature at location $(u,v)$ for sample $i$, and $\alpha$ controls the sharpness of the contrastive margin.

%% file: secs/Experiments.tex
\section{Experiments}

\begin{table}[t]
\centering
\small
\setlength{\abovecaptionskip}{0pt}
\setlength{\belowcaptionskip}{0pt}
\caption{Monocular depth estimation results on the KITTI dataset. All compared methods are evaluated without training on KITTI. Our method, trained solely on nuScenes, achieves the best performance.\label{Tb:KITTI_depth}}
\begin{tabular}{c|ccccccc}
\midrule
                 & $\downarrow$AbsRel         & $\downarrow$SqRel          & $\downarrow\text{RMSE}$  & $\downarrow\text{RMSE}_{log}$ & $\uparrow\delta_1$       & $\uparrow\delta_2$       & $\uparrow\delta_3$ \\ \midrule
DepthPro\cite{bochkovskii2024depth}        & 0.1833          & 0.7400          & 3.6684          & 0.1872          & 0.7361            & 0.9813            & 0.9972            \\
MoGeV2\cite{wang2025moge}          & 0.1822          & 0.6595          & 3.8175          & 0.2191          & 0.6142            & 0.9786            & 0.9970            \\
DA3\cite{lin2025depth} & 0.2263          & 1.0362          & 4.4589          & 0.2198          & 0.5512            & 0.9860            & 0.9964            \\
\textbf{Ours}            & \textbf{0.1371} & \textbf{0.5361} & \textbf{3.4067} & \textbf{0.1558} & \textbf{0.8542}   & \textbf{0.9881}   & \textbf{0.9976}   \\ \midrule
\end{tabular}
\vspace{-1em}
\end{table}

\textbf{Implementation details.}
We follow the protocol used in prior work~\cite{shi2022beyond,xia2023convolutional} to define coarse camera poses. We simulate noisy GPS initialization by perturbing the ground-truth latitude and longitude within $\pm 20\,\mathrm{m}$, while perturbing the yaw angle within $\pm 10^{\circ}$.
We adopt $\pi^{3}$~\cite{wang2025pi} as the representative backbone and extend it with modules for global scale estimation, coordinate-offset regression, and geometric correction. The model is trained for 100 epochs, during which the number of ground-view images $N$ is randomly sampled between 1 and 3. During the first 80 epochs, the global scale and coordinate offset are optimized to establish metric alignment. The remaining 20 epochs are devoted to geometric correction, refining the reconstructed scene geometry based on the estimated scale. This staged training strategy improves optimization stability by resolving scale ambiguity before applying fine-grained geometric refinement. All training and evaluation are conducted on four NVIDIA L40 GPUs.

\begin{figure}[t]
    \centering
    \includegraphics[width=1\textwidth]{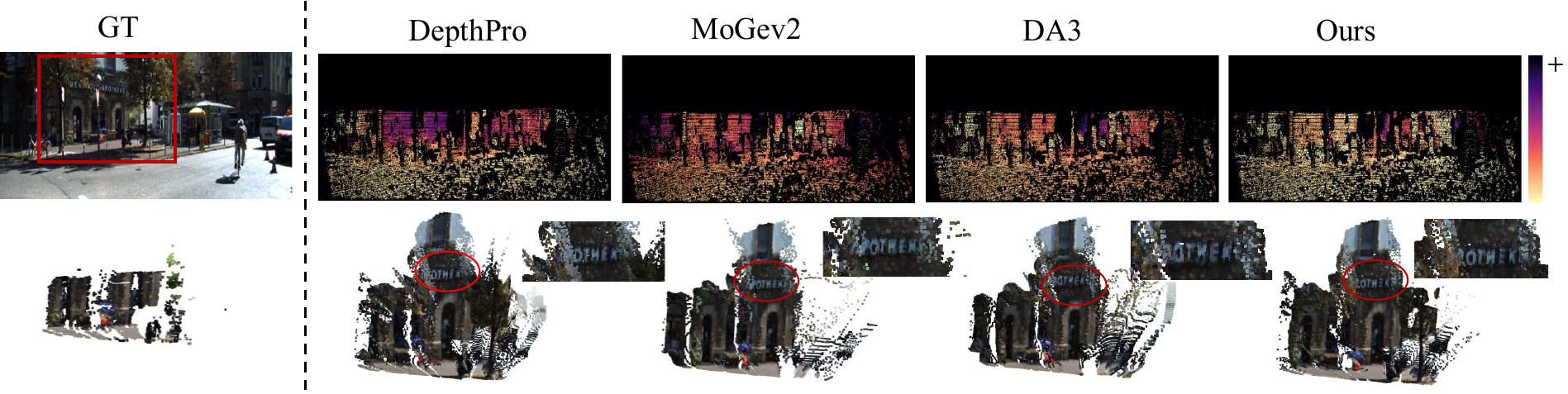}
    \caption{Visualization of monocular metric depth estimation. The top-left shows the ground-truth (GT) image, and below it is the point cloud corresponding to the building enclosed by the red box. The top-right displays the absolute difference between predicted depth and GT depth, with lighter colors indicating smaller errors. The bottom-right shows the predicted point cloud of the building.}
    \label{Fig:monocular}
    \vspace{-1em}
\end{figure}

\begin{table}[t]
\centering
\small
\setlength{\abovecaptionskip}{0pt}
\setlength{\belowcaptionskip}{0pt}
\caption{Multi-view metric depth estimation on the KITTI dataset. In the upper part, all methods are aligned to the ground-truth metric scale. In the lower part, metric-scale predictions are directly compared against ground truth.\label{Tb:KITTI_video_depth}}
\begin{tabular}{c|c|ccccccc}
\midrule
       & Scale                 & $\downarrow$AbsRel         & $\downarrow$SqRel          & $\downarrow\text{RMSE}$  & $\downarrow\text{RMSE}_{log}$ & $\uparrow\delta_1$       & $\uparrow\delta_2$       & $\uparrow\delta_3$       \\ \midrule
Fast3R\cite{yang2025fast3r} & \multirow{6}{*}{GT}   & 0.1024          & 0.5489          & 3.9426          & 0.1501               & 0.9053          & 0.9789          & 0.9935          \\
Mast3R\cite{mast3r_eccv24} &                       & 0.0834          & 0.6435          & 4.6779          & 0.3605               & 0.9445          & 0.9774          & 0.9861          \\
PI3\cite{wang2025pi}    &                       & 0.0464          & 0.1488          & 2.1142          & 0.0748               & 0.9828          & 0.9970          & \textbf{0.9989} \\
VGGT\cite{wang2025vggt}   &                       & 0.0703          & 0.2904          & 3.2424          & 0.1064               & 0.9618          & 0.9899          & 0.9966          \\
DA3\cite{lin2025depth}    &                       & 0.0473          & 0.2601          & 2.5869          & 0.0875               & 0.9754          & 0.9938          & 0.9978          \\
\textbf{Ours}   &                       & \textbf{0.0353} & \textbf{0.1219} & \textbf{2.0039} & \textbf{0.0634}      & \textbf{0.9896} & \textbf{0.9972} & \textbf{0.9989} \\ \midrule
DA3\cite{lin2025depth}     & \multirow{2}{*}{Pred} & 0.2448          & 1.1920          & 5.0979          & 0.3014               & 0.3588          & 0.8582          & 0.9754          \\
\textbf{Ours}   &                       & \textbf{0.1226} & \textbf{0.3605} & \textbf{2.9460} & \textbf{0.1475}      & \textbf{0.8590} & \textbf{0.9988} & \textbf{0.9989} \\ \midrule
\end{tabular}
\vspace{-1em}
\end{table}

\subsection{Scene Geometry Estimation}
To evaluate scene geometry estimation, we conduct experiments on three tasks: monocular metric depth estimation, multi-view metric depth estimation, and multi-view metric point-cloud estimation. Cross-dataset evaluation is performed on KITTI~\cite{geiger2013vision,shi2022beyond} and nuScenes~\cite{nuscenes2019,fong2021panoptic} to assess generalization. Specifically, models trained on KITTI are evaluated on nuScenes, and models trained on nuScenes are evaluated on KITTI.
For KITTI, we use the satellite imagery from~\cite{shi2022beyond}. For nuScenes, we augment the dataset with satellite images to support cross-view supervision. The satellite images are retrieved from Google Maps~\cite{googlemaps_maps} using publicly available GPS annotations, with each satellite image covering an area of $250\,\mathrm{m} \times 250\,\mathrm{m}$.

\textbf{Monocular metric depth estimation.}
We evaluate monocular depth estimation on KITTI. The model is trained exclusively on nuScenes and directly tested on KITTI, yielding a strict cross-dataset generalization setting. As shown in Table~\ref{Tb:KITTI_depth}, we report Absolute Relative Error (AbsRel), Squared Relative Error (SqRel), Root Mean Squared Error ($\text{RMSE}$), Root Mean Squared Logarithmic Error ($\text{RMSE}_{\log}$), and the accuracy thresholds $\delta_1 < 1.25$, $\delta_2 < 1.25^2$, and $\delta_3 < 1.25^3$. Compared with existing monocular metric depth estimation methods~\cite{bochkovskii2024depth,wang2025moge,lin2025depth}, our approach achieves superior performance across most metrics. This improvement comes from incorporating satellite imagery as an explicit metric reference. Instead of relying solely on learned dataset-specific priors to infer absolute scale, our method aligns ground-view images with satellite imagery, enabling more reliable recovery of metric depth and better generalization to unseen environments. As shown in Figure~\ref{Fig:monocular}, the scale estimated by our method is closest to the ground truth, and our reconstruction captures finer geometric details.

\textbf{Multi-view metric depth estimation.}
We further evaluate multi-view metric depth estimation on KITTI. The model is trained solely on nuScenes and evaluated directly on KITTI.
As shown in the lower part of Table~\ref{Tb:KITTI_video_depth}, we compare with DepthAnythingV3~\cite{lin2025depth}, a strong metric multi-frame depth estimation method. Both methods directly produce metric-scale depth maps that are compared against ground truth. Our approach achieves consistent improvements across all metrics, largely due to the additional scale prior provided by satellite imagery. In contrast, DepthAnythingV3 relies more heavily on priors learned from large-scale training data, which can limit its performance on unseen scenes. In the upper part of Table~\ref{Tb:KITTI_video_depth}, we additionally compare with scale-agnostic approaches~\cite{yang2025fast3r,mast3r_eccv24,wang2025vggt,wang2025pi,lin2025depth}. For a fair comparison, their outputs are aligned to the ground-truth metric scale using an optimal scale factor. Even under this favorable setting, our method shows clear advantages, suggesting that satellite-guided geometric correction $\Delta X_N$ improves the initial reconstruction and yields more accurate metric-scale depth estimates.

\begin{figure}[t]
    \centering
    \includegraphics[width=1\textwidth]{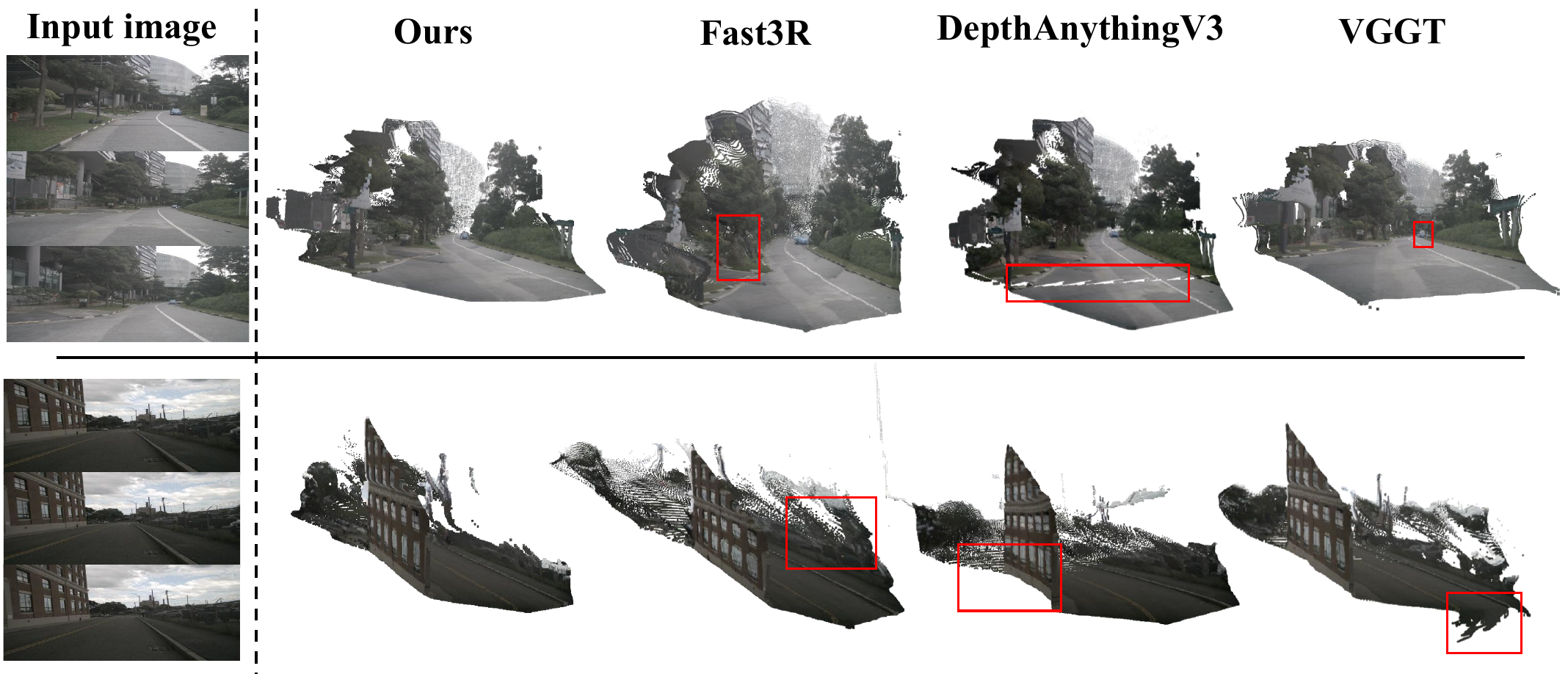}
    \caption{Visualization of multi-view point-cloud estimation. Our method produces reconstructions that are more complete and less noisy than competing approaches.}
    \label{Fig:pointcloud}
    \vspace{-1em}
\end{figure}

\begin{table}[t]
\centering
\small
\setlength{\abovecaptionskip}{0pt}
\setlength{\belowcaptionskip}{0pt}
\caption{Multi-view point-cloud estimation on the nuScenes dataset. In the upper part, all methods are aligned to the ground-truth scale. In the lower part, metric predictions are directly compared with the ground-truth point map.\label{Tb:nuScenes_point}}
\setlength{\tabcolsep}{6pt}
\begin{tabular}{c|c|cc|cc|cc}
\midrule
\multirow{2}{*}{} & \multirow{2}{*}{Scale} & \multicolumn{2}{c|}{$\downarrow$Acc.}         & \multicolumn{2}{c|}{$\downarrow$Comp.}        & \multicolumn{2}{c}{$\uparrow$N.C.}           \\ \cline{3-8} 
                  &                        & Mean            & Med.             & Mean            & Med.             & Mean            & Med.             \\ \midrule
Fast3r\cite{yang2025fast3r}            & \multirow{6}{*}{GT}    & 1.3420          & 0.7876          & 1.9050          & 0.9322          & 0.7495          & 0.8866          \\
Mast3r\cite{mast3r_eccv24}            &                        & 1.4066          & 0.9125          & 1.8220          & 1.1100          & 0.7451          & 0.8741          \\
Pi3\cite{wang2025pi}               &                        & 1.0041          & 0.5590          & 1.3636          & 0.8599          & \textbf{0.8006} & \textbf{0.9442} \\
VGGT\cite{wang2025vggt}              &                        & 1.0736          & 0.6967          & \textbf{1.1242} & \textbf{0.5537} & 0.7964          & 0.9368          \\
DA3\cite{lin2025depth}               &                        & 1.2462          & 0.6762          & 1.4774          & 0.8321          & 0.7841          & 0.9315          \\
\textbf{Ours}              &                        & \textbf{0.9922} & \textbf{0.5566} & 1.3556          & 0.8340          & \textbf{0.8006} & 0.9439          \\ \midrule
DA3\cite{lin2025depth}               & \multirow{2}{*}{Pred}  & 1.3637          & \textbf{0.8045} & 2.0295          & 1.0040          & 0.7765          & 0.9223          \\
\textbf{Ours}              &                        & \textbf{1.3632} & 0.8249          & \textbf{1.4166} & \textbf{0.8471} & \textbf{0.7846} & \textbf{0.9317} \\ \midrule
\end{tabular}
\vspace{-1em}
\end{table}

\textbf{Multi-view point-cloud estimation.}
We further evaluate multi-view point-cloud estimation by applying the model trained on KITTI directly to nuScenes. We report Accuracy (Acc.), Completion (Comp.), and Normal Consistency (N.C.) in Table~\ref{Tb:nuScenes_point}. In the lower part of the table, both our method and DepthAnythingV3~\cite{lin2025depth} predict metric geometry and are directly compared with the ground-truth point cloud without scale alignment. In the upper part, we compare with scale-agnostic approaches~\cite{yang2025fast3r,mast3r_eccv24,wang2025vggt,wang2025pi,lin2025depth}, whose outputs are aligned to the ground-truth metric scale using a single scale factor. Our method achieves superior performance on most metrics, showing that satellite imagery improves geometric reconstruction quality. As shown in Figure~\ref{Fig:pointcloud}, our reconstructions are more accurate and complete, producing denser point clouds with fewer outliers.

\subsection{Metric Pose Estimation}

\begin{figure}[t]
    \centering
    \includegraphics[width=1\textwidth]{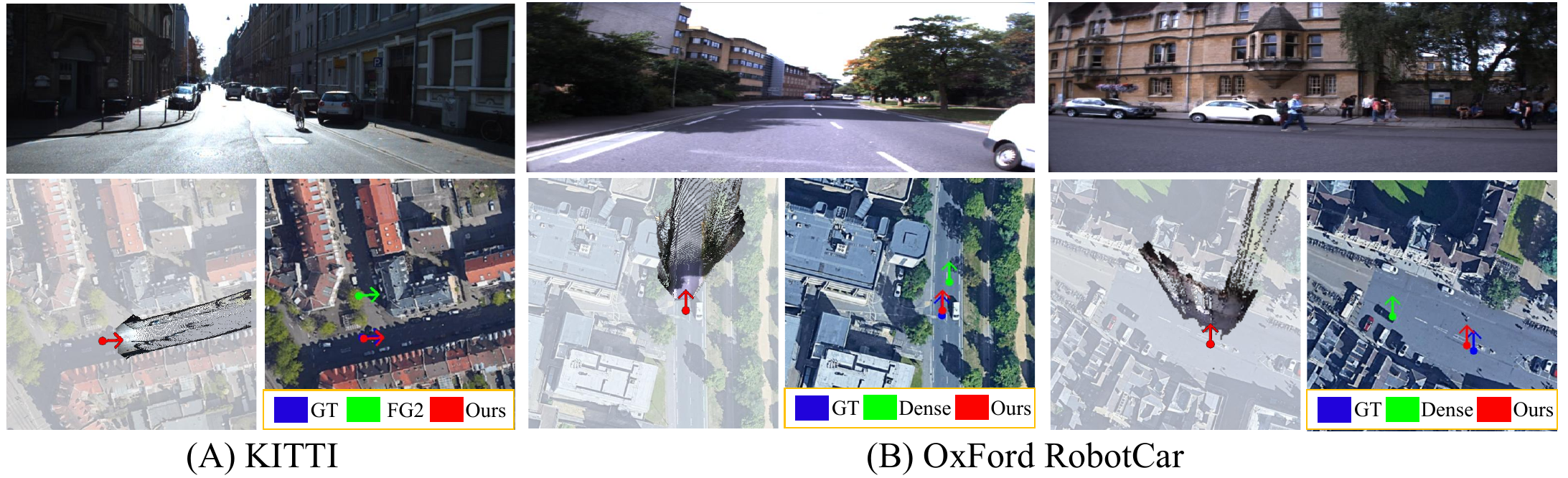}
    \caption{Visualization of localization results on the KITTI and Oxford RobotCar datasets. In each group, the top image shows the ground-view image; the bottom-left image projects the reconstructed ground scene onto the satellite image; and the bottom-right image indicates the estimated camera position within the satellite map.}
    \label{Fig:localization}
    \vspace{-1em}
\end{figure}

\begin{table}[t]
\centering
\small
\setlength{\abovecaptionskip}{0pt}
\setlength{\belowcaptionskip}{0pt}
\caption{Location and orientation estimation error on KITTI. The best results are highlighted in bold, and the second-best results are underlined.\label{Tb:KITTI_localization}}
\setlength{\tabcolsep}{6pt}
\begin{tabular}{c|cccc|cccc}
\midrule
          & \multicolumn{4}{c|}{Same Area}                                                                  & \multicolumn{4}{c}{Cross Area}                                                                 \\ \midrule
          & \multicolumn{2}{c|}{$\downarrow$ Loc. (m)}                    & \multicolumn{2}{c|}{$\downarrow$ Orien. ($^{\circ}$)} & \multicolumn{2}{c|}{$\downarrow$ Loc. (m)}                    & \multicolumn{2}{c}{$\downarrow$ Orien. ($^{\circ}$)} \\
          & Mean          & \multicolumn{1}{c|}{Med.}          & Mean                 & Med.                & Mean          & \multicolumn{1}{c|}{Med.}          & Mean                & Med.                \\ \midrule
CCVPE\cite{xia2023convolutional}     & 1.22          & \multicolumn{1}{c|}{0.62}          & 0.67                 & 0.54                & 9.16          & \multicolumn{1}{c|}{3.33}          & \underline{1.55}                & \underline{0.84}                \\
DenseFlow\cite{song2023learning} & 1.48          & \multicolumn{1}{c|}{\textbf{0.47}} & \underline{0.49}                 & \textbf{0.30}       & 7.97          & \multicolumn{1}{c|}{3.52}          & 2.17                & 1.21                \\
FG2\cite{xia2025fg}       & \underline{0.75}          & \multicolumn{1}{c|}{0.52}          & 1.28                 & 0.74                & 7.45          & \multicolumn{1}{c|}{4.03}          & 3.33                & 1.88                \\
BEVSplat\cite{wang2025bevsplat}  & 2.87          & \multicolumn{1}{c|}{2.06}          & -                    & -                   & \underline{6.20}          & \multicolumn{1}{c|}{\underline{2.51}}          & -                   & -                   \\
\textbf{Ours}      & \textbf{0.73} & \multicolumn{1}{c|}{\underline{0.49}}          & \textbf{0.43}        & \underline{0.38}                & \textbf{5.75} & \multicolumn{1}{c|}{\textbf{2.34}} & \textbf{0.52}       & \textbf{0.37}       \\ \midrule
\end{tabular}
\vspace{-1em}
\end{table}

Our method estimates absolute camera pose in the world coordinate system by aligning the reconstructed scene with satellite imagery. The most closely related task is cross-view localization~\cite{shi2022beyond,shi2023boosting,xia2025fg,song2023learning}, where satellite images are used to determine the geographic position of ground-level images.
We conduct localization experiments on Cross-View KITTI~\cite{shi2022beyond} and Cross-View Oxford RobotCar~\cite{xia2022visual}, both of which provide paired ground-level and satellite imagery. Cross-View KITTI contains one training split and two test splits: Test1 samples images from the same geographic region as the training set, while Test2 uses different regions to evaluate geographic generalization. Cross-View Oxford RobotCar contains one training split, one validation split, and three test splits corresponding to traversals captured on dates different from the training set, enabling evaluation under temporal variation.

\begin{table}[t]
\centering
\small
\setlength{\abovecaptionskip}{0pt}
\setlength{\belowcaptionskip}{0pt}
\caption{Location estimation errors on the Oxford RobotCar dataset. As in previous work, the orientation of the ground image is assumed to be known.\label{Tb:Oxford_localization}}
\setlength{\tabcolsep}{4pt}
\begin{tabular}{c|cc|cc|cc|cc}
\midrule
                                                  & \multicolumn{2}{c|}{Test1}    & \multicolumn{2}{c|}{Test2}    & \multicolumn{2}{c|}{Test3}    & \multicolumn{2}{c}{Overall}   \\
                                                  & $\downarrow$Mean         & $\downarrow$Med.         & $\downarrow$Mean         & $\downarrow$Med.         & $\downarrow$Mean         & $\downarrow$Med.         & $\downarrow$Mean         & $\downarrow$Med.         \\ \midrule
MCC\cite{xia2022visual}          & 1.42          & 1.10          & 1.95          & 1.33          & 1.94          & 1.29          & 1.77          & 1.24          \\
CVR\cite{zhu2021vigor}                                               & 1.88          & 1.47          & 2.64          & 1.99          & 2.35          & 1.71          & 2.29          & 1.72          \\
DenseFlow\cite{song2023learning} & 1.17          & \textbf{0.72} & 1.76          & 0.97          & 1.79          & 0.92          & 1.57          & 0.87          \\
Boosting\cite{shi2023boosting}   & 2.40          & 0.91          & 3.10          & 1.13          & 2.86          & 1.05          & 2.79          & 1.02          \\
\textbf{Ours}                                              & \textbf{1.02} & 0.87          & \textbf{1.09} & \textbf{0.90} & \textbf{1.08} & \textbf{0.74} & \textbf{1.07} & \textbf{0.84} \\ \midrule
\end{tabular}
\vspace{-1em}
\end{table}

\begin{table}[t]
\centering
\small
\setlength{\abovecaptionskip}{0pt}
\setlength{\belowcaptionskip}{0pt}
\caption{Ablation study on the nuScenes dataset.\label{Tb:ablation}}
\setlength{\tabcolsep}{5.5pt}
\begin{tabular}{c|cc|cc|cc}
\midrule
\multirow{2}{*}{} & \multicolumn{2}{c|}{$\downarrow$Acc.}         & \multicolumn{2}{c|}{$\downarrow$Comp.}        & \multicolumn{2}{c}{$\uparrow$N.C.}           \\ \cline{2-7} 
                  & Mean            & Med.             & Mean            & Med.             & Mean            & Med.             \\ \midrule
w/o Sat.          & 1.8701          & 1.425          & 4.699         & 1.675         & 0.7162          & 0.8643          \\
w/o Correction     & 1.9894          & 1.1060          & 1.9730          & 1.4358          & 0.7639          & 0.9172          \\
\textbf{Ours}              & \textbf{1.3632} & \textbf{0.8249} & \textbf{1.4166} & \textbf{0.8471} & \textbf{0.7846} & \textbf{0.9317} \\ \midrule
\end{tabular}
\vspace{-1em}
\end{table}

Following prior cross-view localization protocols, we simulate noisy GPS initialization by perturbing the ground-truth pose with random offsets of $\pm 20\,\mathrm{m}$ in latitude and longitude and $\pm 10^{\circ}$ in yaw. We evaluate camera localization using position error and yaw error, reporting both mean and median Euclidean localization errors. Across both datasets, our method consistently outperforms all baselines by a significant margin. The particularly strong performance on the cross-area split of KITTI highlights its ability to generalize across unseen geographic regions. As shown in Figure~\ref{Fig:localization}, projecting the reconstructed ground scene onto the satellite image produces precise geometric alignment, indicating that our method effectively recovers metric scale and improves absolute pose estimation.

\subsection{Ablation Study}
To validate the effectiveness of each component, we perform a systematic ablation study by selectively removing key modules. All experiments train on KITTI and evaluate multi-view point-map estimation on nuScenes, testing cross-dataset generalization.
First, we remove the satellite reference and train the model using only ground-view inputs. In this setting, the model learns only a limited notion of scale from the training data, and the estimated scale remains unstable in unseen environments. Next, we evaluate the geometric correction module. By leveraging satellite imagery as an external reference, this module refines the initial point-map predictions, correcting residual misalignment and improving completeness.